%% file: MSCameraAutomation_qlin.tex
\documentclass[english]{ieeeconf}
\usepackage[T1]{fontenc}
\usepackage[latin9]{inputenc}
\usepackage[letterpaper]{geometry}
\usepackage{color}
\usepackage{array}
\usepackage{textcomp}
\usepackage{url}
\usepackage{bm}
\usepackage{amsmath}
\usepackage{amssymb}
\usepackage{graphicx}
\usepackage{setspace}

\makeatletter

\newcommand{\lyxmathsym}[1]{\ifmmode\begingroup\def\b@ld{bold}
  \text{\ifx\math@version\b@ld\bfseries\fi#1}\endgroup\else#1\fi}

\providecommand{\tabularnewline}{\\}

\IEEEoverridecommandlockouts
\usepackage{resizegather}
\usepackage{breqn}
\makeatother

\usepackage{babel}
\begin{document}
\input{macros/macros.tex}

\input{macros/macros_optimization.tex}
\input{macros/macros_adaptive.tex}

\title{Autonomous Field-of-View Adjustment Using Adaptive Kinematic Constrained
Control with Robot-Held Microscopic Camera Feedback}
\author{Hung-Ching Lin, Murilo Marques Marinho, Kanako Harada\thanks{This
work was supported by JST Moonshot R\&D JPMJMS2033.}\thanks{Hung-Ching
Lin and Kanako Harada are with the Department of Mechanical Engineering,
the University of Tokyo, Tokyo, Japan. \texttt{Emails:}{\texttt{qlin1806,
kanako}}\texttt{@g.ecc.u-tokyo.ac.jp}. } \thanks{Murilo M. Marinho
is with the Department of Electrical and Electronic Engineering, the
University of Manchester, Manchester, UK. \texttt{Email:}murilo.marinho\texttt{@}manchester.ac.uk}\vspace{-1em}}
\maketitle
\begin{abstract}
Robotic systems for manipulation in millimeter scale often use a camera
with high magnification for visual feedback of the target region.
However, the limited field-of-view (FoV) of the microscopic camera
necessitates camera motion to capture a broader workspace environment.
In this work, we propose an autonomous robotic control method to constrain
a robot-held camera within a designated FoV. Furthermore, we model
the camera extrinsics as part of the kinematic model and use camera
measurements coupled with a U-Net based tool tracking to adapt the
complete robotic model during task execution. As a proof-of-concept
demonstration, the proposed framework was evaluated in a bi-manual
setup, where the microscopic camera was controlled to view a tool
moving in a pre-defined trajectory. The proposed method allowed the
camera to stay 94.1\% of the time within the real FoV, compared to
54.4\% without the proposed adaptive control.\thispagestyle{empty}
\end{abstract}

\section{Introduction}

\setlength{\textfloatsep}{0pt}
\setlength{\intextsep}{0pt}
\setlength{\abovecaptionskip}{0pt}
\setlength{\parindent}{0pt} 

One major hurdle in robotic systems for manipulation tasks under microscopic
view is providing situational awareness to operators or autonomous
agents to close the task control loop effectively. A single fixed
(or manually moved) high-magnification camera vision system is a common
solution to this challenge. At this scale, due to the limited resolution
of image sensors for digital zoom, the limited field-of-view of the
camera, and the restricted working distance of lenses with high magnification,
constant motion of the camera is often needed to capture essential
points-of-view of the environment.

One pertinent example is \textit{intravital imaging} \cite{koike_modelling_2019}.
This procedure involves installing a transparent observatory cranial
window in mouse skulls with an 8 mm glass, enabling scientists to
monitor human organoid growth in the mouse brain \emph{in-vivo}. We
have developed a multi-arm robotic platform \cite{marinho_design_2022}
for performing scientific exploration experiments, showing that we
can perform cranial window drilling in mock egg-shell trials and \emph{ex-vivo}
mice through teleoperation \cite{marinho_design_2022} and autonomously
\cite{zhao_autonomous_2023}. Thanks to the progress of these earlier
works, we can explore the next step of the procedure, such as organoid
implantation. This would demand a higher magnification camera with
a shorter focus distance range (approximately $\pm$5 mm) and a smaller
field of view (approximately 1 cm X 1 cm). Therefore, the task would
effectively require camera motion to maintain the strict \textasciitilde 40cm
focus distance while operating in the larger workspace. 
\begin{figure}
\begin{singlespace}
\centering{}\includegraphics[viewport=0bp 0bp 6004bp 4002bp,clip,width=1\columnwidth]{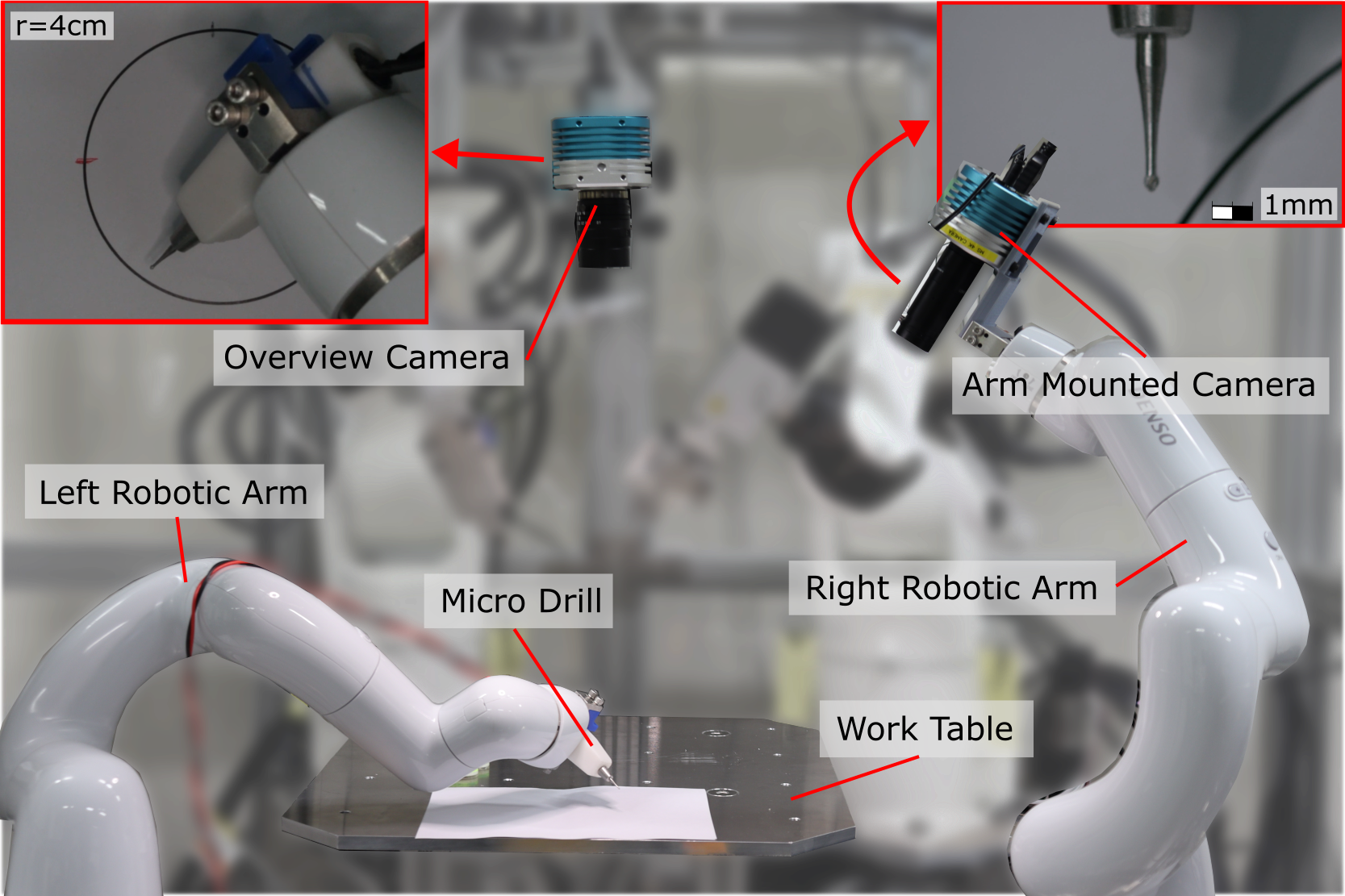}\caption{\label{fig:sytem}The system configuration presented in this paper
comprises two of the four arms from our AI-Robot Platform for Scientific
Exploration \cite{marinho_design_2022}. On the left is an arm holding
a drill. On the right is an arm holding a camera. Above the workspace
is a overview camera that provides stationary view of the workspace.}
\end{singlespace}
\end{figure}

In this context, we hypothesize that a viable solution is to mount
a camera on the robotic arm, making it controllable. This approach
attempts to ensure that the field of view remains focused on the tool,
whether controlled via teleoperation or an autonomous agent. The concept
is inspired by existing literature on \emph{minimally invasive surgery}
(MIS) in topics such as endoscope automation. In MIS systems, of which
the da Vinci Surgical System \cite{guthart_intuitivesup_2000} is
an example, the robotic endoscope is controllable through teleoperation.

A major difficulty in this context is that a constrained workspace
shared by many arms must have model-based self-collision avoidance
\cite{marinho_design_2022}. The model comprises parameters related
to the geometry of the robots and, additionally, in this work, the
\emph{extrinsics} of the mounted camera. In our target environment,
owing to size constraints, an \emph{offline} calibration procedure
using additional sensors with large footprints, such as \cite{gharaaty_online_2018}
and \cite{yu_simultaneous_2018}, is not viable. Instead, we are interested
in exploiting camera information to calibrate the model \emph{online},
through adaptive control \cite{marinho_adaptive_2022}.

\subsection{\textit{Related works}}

\begin{table*}[t]
\caption{\label{tab:related_works}Capabilities of the proposed work in contrast
with existing literature.}

\begin{centering}
\scalebox{0.8}{%
\begin{tabular*}{2\columnwidth}{@{\extracolsep{\fill}}|>{\centering}m{3cm}|>{\centering}m{1.125cm}||>{\centering}m{0.9cm}|>{\centering}m{0.9cm}|>{\centering}m{0.9cm}|>{\centering}m{0.9cm}|>{\centering}m{0.9cm}|>{\centering}m{0.9cm}|>{\centering}m{0.9cm}|>{\centering}m{0.9cm}|>{\centering}m{0.9cm}|}
\hline 
 & This work & \cite{koyama_autonomous_2022} & \cite{hsu_adaptive_1999} & \cite{zergeroglu_vision-based_2001} & \cite{cheah_adaptive_2006} & \cite{cheah_adaptive_2010} & \cite{zhong_robots_2015} & \cite{wang_vision-based_2018}\textbf{{*}} & \cite{bechlioulis_robust_2019} & \cite{miao_vision-based_2021}\tabularnewline
\hline 
\hline 
Visual servoing & $\checkmark$ & $\checkmark$ & $\checkmark$ & $\checkmark$ & $\checkmark$ & $\checkmark$ & $\checkmark$ & $\times$ & $\checkmark$ & $\checkmark$\tabularnewline
\hline 
Marker-less & $\checkmark$ & $\checkmark$ & $\times$ & $\times$ & $\times$ & $\times$ & $\checkmark$ & $\triangle$ & $\times$ & $\times$\tabularnewline
\hline 
Adaptive (kinematics) & $\checkmark$ & $\times$ & $\checkmark$ & $\checkmark$ & $\checkmark$ & $\checkmark$ & $\times$ & $\checkmark$ & $\times$ & $\checkmark$\tabularnewline
\hline 
Adaptive (extrinsics) & $\checkmark$ & $\times$ & $\times$ & $\times$ & $\checkmark$ & $\times$ & $\checkmark$ & $\times$ & $\times$ & $\times$\tabularnewline
\hline 
Workspace constraints & $\checkmark$ & $\checkmark$ & $\times$ & $\times$ & $\times$ & $\checkmark$ & $\times$ & $\times$ & $\times$ & $\times$\tabularnewline
\hline 
FoV constraints & $\checkmark$ & $\checkmark$ & $\times$ & $\times$ & $\times$ & $\times$ & $\checkmark$ & $\times$ & $\checkmark$ & $\checkmark$\tabularnewline
\hline 
Scale & mm & mm & \textcolor{black}{cm} & \textcolor{black}{cm} & cm & m & cm & mm & m & m\tabularnewline
\hline 
\end{tabular*}}
\par\end{centering}

\medskip{}

\centering{}\textbf{\small{}{*}}{\small{} In \cite{wang_vision-based_2018},
marker-based tracking was used for obtaining ground-truth.}\vspace*{-15pt}
\end{table*}

A significant portion of the related work on camera automation is
in the domain of MIS systems. For instance, Ali \textit{et al.} \cite{ali_eye_2008}
and Dardona \textit{et al.} \cite{dardona_remote_2019} focused on
alleviating the surgeon's workload through novel methods for camera
control input using eye gaze tracking and head-mounted display, respectively.
In works toward the automation of the camera movement, King \textit{et
al.} \cite{king_towards_2013} employed color markers on effectors
to track the tools in a surgical procedure, subsequently using this
data to generate the control input for endoscopes based on a predefined
set of rules. Extending \cite{king_towards_2013}, Eslamian \textit{et
al. }\cite{eslamian_development_2020} implemented the method on the
da Vinci Surgical System for comparison against the traditional clutch-based
camera control. In a different approach, Ji\textit{ et al.} \cite{ji_learning_2018}
proposed a method of identifying viewpoints of interest in a camera
view to automatically re-center the camera learning from surgeons'
demonstrations. In more related examples that also perform adaptation
of robot parameters, Wang \textit{et al.} \cite{wang_vision-based_2018}
proposed a method that makes use of the property of the RCM-based
robot for online calibration with an endoscopic camera end joint encoder.
Outside of the MIS domain, early work by Hsu \textit{et al.} \cite{hsu_adaptive_1999}
and Zergeroglu \textit{et al.} \cite{zergeroglu_vision-based_2001}
have both proposed methods for adaptive control of fixed cameras on
planer robots. Later work by Cheah \textit{et al.} in \cite{cheah_adaptive_2006}
also proposed a method for visual servoing using the shadow cast by
the end-effector for parameter adaptation, and in \cite{cheah_adaptive_2010},
was extended to include force information and allow for tracking on
a workspace constraint surface.

Regarding camera automation works that also took into account the
FoV constraints, Chesi and Hung \cite{chesi_global_2007} proposed
a method for path planning accounting for workspace constraints in
addition to FoV constraints, and Bechlioulis \textit{et al.} \cite{bechlioulis_robust_2019}
demonstrated a visual servoing that is robust to depth measurement
error. In a slightly different approach, the work by Zhong \textit{et
al.} \cite{zhong_robots_2015}proposed a method of servoing without
an explicit model of the robot and using a neural network to assist
in state estimation.

We have shown in our prior work by Koyama \textit{et al.} \cite{koyama_autonomous_2022}
that it is possible to effectively use the robot's model to automate
the light-guide in an automated bimanual setup to provide consistent
lighting conditions. Image tracking of the tooltip was used to move
the robot to a set point but with no adaptation of the robot parameters.
However, in such scenario, even minor discrepancies in robot parameters
or change in effector weight and configuration often result in relatively
large misalignment in the pose of the robot effector. Gonzales \textit{et
al.} \cite{gonzalez_online_2022} proposed a method for compliance
compensation correcting for errors from weight or load on the robot
and is strong evidence that robot misalignment can be difficult to
model and impossible to calibrate offline for all trajectories. Marinho
and Adorno \cite{marinho_adaptive_2022} have recently introduced
an adaptive constrained kinematic control that can utilize partial
sensor information. In this work, our interest is to control the instrument
with FoV constraints while the camera tracking of the tooltip is used
to adjust the robot model along with the camera extrinsics. Capabilities
of the works described herein are shown in Table~\ref{tab:related_works}.

\subsection{\textit{Statement of contributions}}

In this work, we propose a FoV camera automation framework with parameter
adaptation by (1) modeling the problem with a proper partial measurement
space definition, (2) obtaining the camera extrinsics Jacobian and
projector to use it in an adaptive control law; and (3) assessing
our system on tracking drill position in a proof-of-concept scenario.

\section{Problem statement}

Consider the system in Fig. \ref{fig:sytem} as part of our robot
platform for scientific exploration \cite{marinho_design_2022}, from
which we use two of the 8 DoF robotic branches, each one composed
of a motor to rotate about the center of the workspace, a linear actuator
to move along the radius of the rail, and a robotic arm (CVR038, Densowave,
Japan). Let the first robot branch, $R1$, have configuration space
$\myvec q_{1}\in\mathbb{R}^{8}$ and parameter space $\estimated{\myvec a}_{1}\in\mathbb{R}^{44}$
holding a micro drill (MD1200, Braintree Scientific, USA). Let the
second robot branch, $R2$, be the robotic arm (CVR038, Densowave,
Japan) with joint values $\myvec q_{2}\in\mathbb{R}^{8}$ and parameter
space $\estimated{\myvec a}_{2}\in\mathbb{R}^{44}$ holding a robot-mounted
camera (STC-HD853HDMI, Omron-Sentech, Japan) with microscopic lenses
(VS-LDA75, VS Technology, Japan). The robotic system as a whole can
be seen as having configuration space $\myvec q=\begin{bmatrix}\myvec q_{1}^{T} & \myvec q_{2}^{T}\end{bmatrix}^{T}$
and parameter space $\estimated{\myvec a}=\begin{bmatrix}\estimated{\myvec a}_{1}^{T} & \estimated{\myvec a}_{2}^{T}\end{bmatrix}^{T}$.

The image of the micro drill's workspace is obtained through the $R2$'s
robot-mounted camera to provide the operator with a closed-up view
of the task. To it, we add a U-Net \cite{ronneberger_u-net_2015}
based keypoint detection tracking for the tip of the drill. The tracker
outputs the center-adjusted pixel coordinates of the drill tip, i.e.
$\quat{\rho}^{\text{oc}}\left(u,v\right)\in\mathbb{I}^{2}$.\vspace*{-7pt}

\subsection{Goal\label{subsec:Goal}}

In this work, we address a centralized kinematic controller that allows
controlling $R1$ to follow a given trajectory $x_{1}\left(t\right)$
in the workspace while autonomously keeping the camera's FoV constraints
\cite{koyama_autonomous_2022} of the camera held by $R2$, and adapting
the robot's parameters, including the camera extrinsics, using the
tooltip information $\left(u,v\right)$.
\begin{figure*}[t]
\begin{singlespace}
\begin{centering}
\vspace*{0pt}
\par\end{centering}
\centering{}\includegraphics[width=0.83\paperwidth]{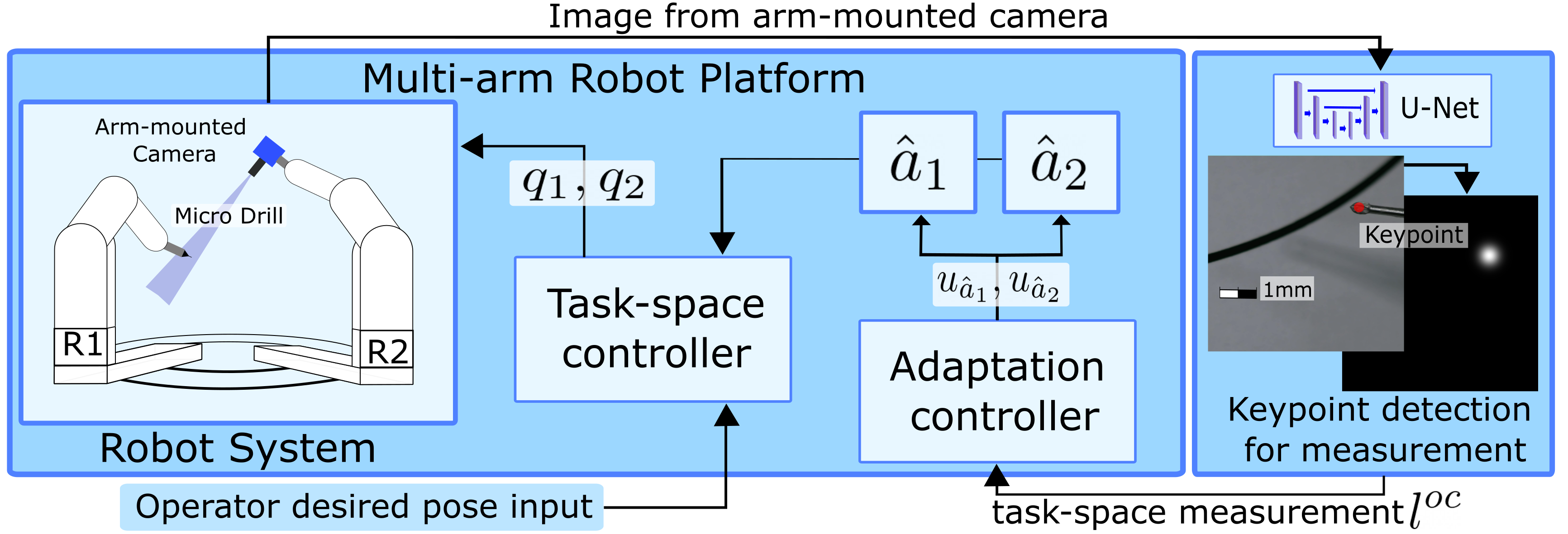}\caption{\label{fig:overview}Block diagram of the proposed system. The task-space
controller receivesthe desired pose control signal to move R1 in a
given trajectory, and both $R1$ and $R2$ to maintain the task constraints,
including the FoV constraint. Meanwhile, a U-Net based tracking algorithm
using images of the robot-mounted camera in $R2$ outputs the tooltip
of the tool held by $R1$. Using these measurements, the adaptive
controller updates the estimated parameters of the kinematic model
of both robots, which include the camera extrinsics. That updated
model is used by the task-space controller, and the task--adaptive
loop is closed.}
\vspace*{-15pt}
\end{singlespace}
\end{figure*}

\section{Mathematical background}

For the control and adaptation models we use (dual) quaternion algebra
based on the notation of \cite{adorno2017robot,marinho_dynamic_2019,marinho_adaptive_2022}.
In a brief explanation for the notation used, a pose is represented
by a unit dual quaternion, e.g. $\dq{\mathcal{S}}\ni\dq x=\quat r+0.5\dual\quat t\quat r$,
where $\dual\neq0$ but $\dual^{2}=0$. The translation is defined
a the pure quaternion, e.g. $\mathbb{H}_{p}\ni\quat t=t_{x}\imi+t_{y}\imj+t_{z}\imk$,
where $\imi^{2}=\imj^{2}=\imk^{2}=\imi\imj\imk=-1$, and can be mapped
into a vector with $\vector_{3}\quat t=\left[\begin{array}{ccc}
t_{x} & t_{y} & t_{z}\end{array}\right]^{T}\in\mathbb{R}^{3}$ to represent a point in space. The rotation is defined by a unit
quaternion, e.g. $\mathbb{S}^{3}\ni\quat r=\cos\left(\phi/2\right)+\quat v\sin\left(\phi/2\right)$,
that represents a rotation of $\phi$ radians about the unit vector
$\quat v\in\mathbb{H}_{p}\cap\mathbb{S}^{3}$ and can be mapped into
a vector with $\vector_{4}\quat r=\left[\begin{array}{cccc}
\cos\left(\phi/2\right) & v_{x}\sin\left(\phi/2\right) & v_{y}\sin\left(\phi/2\right) & v_{z}\sin\left(\phi/2\right)\end{array}\right]^{T}\in\mathbb{R}^{4}$. A Plucker line is a pure dual quaternion with unit norm, e.g. $\mathcal{H}_{p}\cap\dq{\mathcal{S}}\ni\dualvector l=\quat l+\dual\left(\quat p_{l}\times\quat l\right)$,
where $\quat l\in\mathbb{H}_{p}\cap\mathbb{S}^{3}$ is the line direction
and $\quat p_{l}\in\mathbb{H}_{p}$ is a point in the line. Any $\quat v\in\mathbb{H}_{p}\cap\mathbb{S}^{3}$
can be rotated by $\quat r$ with the adjoint, namely $\ad{\quat r}{\quat l}=\quat r\quat l\quat r^{*}$,
where $\quat r^{*}=\cos\left(\phi/2\right)-\quat v\sin\left(\phi/2\right)$,
and $\vector_{4}\quat r\quat l\quat r^{*}=\hami -_{4}\left(\quat r\right)\hami +_{4}\left(\quat r^{*}\right)\vector_{4}\quat l$.

If $\dot{\myvec q}\in\mathbb{R}^{n}$ are the configuration velocities
of a robotic system and $\estimated{\myvec a}\in\mathbb{R}^{p}$ are
the \emph{estimated} kinematic parameters \cite{marinho_adaptive_2022},
then we can define the \emph{estimated} rotation and translation Jacobians
with $\vector_{4}\dot{\estimated{\quat r}}=\mymatrix J_{\estimated r,q}\dot{\myvec q}+\mymatrix J_{\estimated r,\estimated a}\dot{\estimated{\myvec a}}$
and $\vector_{4}\dot{\estimated{\quat t}}=\mymatrix J_{\estimated t,q}\dot{\myvec q}+\mymatrix J_{\estimated t,\estimated a}\dot{\estimated{\myvec a}}$,
respectively.

\section{Proposed camera automation overview}

The overall framework is illustrated in Fig. \ref{fig:overview}.
The task-space controller receives the desired pose signal from either
an operator or predefined trajectory and works to minimize task-space
error of $R1$ using current robot parameters, while both $R1$ and
$R2$ maintain the task constraints. The adaptive controller receives
measurements of an image-processing algorithm to update the estimated
robot parameters to reduce measurement error. The position of the
tooltip of $R1$ is measured using the camera image from $R2$'s robot-mounted
camera processed through a U-Net-based pipeline.

\subsection{U-Net-based point tracking\label{subsec:U-Net-based-point-tracking}}

A U-Net-based \cite{ronneberger_u-net_2015} keypoint detection algorithm
was implemented with \textit{segmentation\_models.pytorch}\footnote{\textit{\url{https://github.com/qubvel/segmentation_models.pytorch}}}
on images down-sampled to $576\times576$ pixels to output a confidence
map for the tip of the drill held by $R1$. We use U-Net as it is
a well established technique with an acceptable level of robustness.
The network used \textit{resnet18} as the encoder initialized with
\textit{ImageNet}\footnote{\textit{\url{https://www.image-net.org/}}}
weight. A total of 2500 original and 17500 augmented images were provided
with common augmentation techniques (e.g., blurs, rescaling, rotate,
color shift, random gamma, brightness and contrast.) , with 9:1 train-validation
split without a testing dataset for 30 epochs in training. Inference
was then performed on a similarly down-sampled camera stream in real-time
to obtain the drill tip pixel coordinate measurement $\quat{\rho}^{\text{oc}}\left(u,v\right)\in\mathbb{I}^{2}$.\vspace*{-7pt}

\subsection{Task controller\label{subsec:Task-controller}}

The task controller is based on \cite{marinho_unified_2018,marinho_dynamic_2019}
with FoV constraints based on \cite{koyama_autonomous_2022}. Let
$\dq{\mathcal{S}}\ni\estimated{\dq x}_{i}=\estimated{\quat r}_{i}+0.5\dual\estimated{\quat t}_{i}\estimated{\quat r}_{i}$
and $\dq{\mathcal{S}}\ni\estimated{\dq x}_{i,d}=\estimated{\quat r}_{i,d}+0.5\dual\estimated{\quat t}_{i,d}\estimated{\quat r}_{i,d}$
be, respectively, the estimated effector poses and desired effector
poses of the two robots $Ri$ with $i\in\left\{ 1,2\right\} $. We
solve, at each time step to obtain a joint velocity control signal
$\myvec u_{q}=\left[\myvec u_{q_{1}}^{T}\,\myvec u_{q_{2}}^{T}\right]^{T}$,
the following optimization problem
\begin{alignat}{1}
\myvec u_{q}\in\underset{\dot{\myvec q}}{\:\text{argmin}}\: & \beta\left(\alpha f_{t,1}+\left(1-\alpha\right)f_{r,1}+\norm{\lambda\dot{\myvec q}_{1}}_{2}^{2}\right)\label{eq:control_law}\\
 & +\left(1-\beta\right)\left(f_{t,2}+\norm{\lambda\dot{\myvec q}_{2}}_{2}^{2}\right)\nonumber \\
\text{subject to}\: & \begin{alignedat}{1}\begin{array}{c}
\mymatrix W_{q}\end{array}\dot{\myvec q} & \preceq\text{\ensuremath{\begin{array}{c}
\myvec w_{q}\end{array}}}\\
\mymatrix B_{q}\dot{\myvec q} & \preceq\myvec b_{q}
\end{alignedat}
\nonumber 
\end{alignat}
in which $\lambda>0\in\mathbb{R}$ is the damping term, $f_{t,i}\triangleq\norm{\mymatrix J_{\estimated t,q_{i}}\dot{\myvec q}_{i}+\eta_{q}\vector_{4}\left(\error{\quat t}_{i}\right)}_{2}^{2}$
is the cost function of the translation error $\error{\quat t}\triangleq\estimated{\quat t}_{i}-\quat t_{i,d}$.
In addition, $f_{r,i}\triangleq\norm{\mymatrix J_{\estimated r,q_{i}}\dot{\myvec q}_{i}+\eta_{q}\vector_{4}\left(\error{\quat r}_{i}\right)}_{2}^{2}$
is the cost function of the switching rotational error
\[
\error{\quat r_{i}}\triangleq\begin{cases}
\left(\estimated{\quat r}_{i}\right)^{*}\quat r_{i,d}-1 & \text{if }\norm{\estimated{\quat r}_{i}^{*}\quat r_{i,d}-1}_{2}<\norm{\estimated{\quat r}_{i}^{*}\quat r_{i,d}+1}_{2}\\
\left(\estimated{\quat r}_{i}\right)^{*}\quat r_{i,d}+1 & \text{otherwise.}
\end{cases}
\]
 The term $\eta_{q}\in\left(0,\infty\right)\subset\mathbb{R}$ is
the proportional task error gain. The term $\alpha\in\left[0,1\right]\subset\mathbb{R}$
is a weight factor for the rotation and translation cost functions.
The term $\beta\in\left[0,1\right]\subset\mathbb{R}$ is weight factor
for the cost functions of each robot, in practice a ``soft'' prioritization
of robot with larger weight. For this work we chose $\eta_{q}=3$,
$\alpha=0.99$, and $\beta=0.999$.

$R1$ is controlled by the desired task-space values $\quat t_{1,d}$
and $\quat r_{1,d}$, whereas $R2$ is autonomously moved by the optimization
problem, with a small preference towards a neutral point-of-view given
by $\quat t_{2,d}$.

\subsubsection{Constraints}

The inequality constraints have two components. The first, obtained
with $\left(\mymatrix W_{q},\begin{array}{c}
\myvec w_{q}\end{array}\right)$ , concerns joint-space constraints, namely position/velocity limits
\cite[Constraint (11)]{marinho_adaptive_2022}. The other component
obtained with $\left(\mymatrix B_{q},\begin{array}{c}
\myvec b_{q}\end{array}\right)$, is related to task-space constraints made of four parts. First,
the collision avoidance of each robot with the environment implemented
with point-to-plane and point-to-line distance constraints \cite[Constraint (59), (32)]{marinho_dynamic_2019},
using 4 points positioned on each robot and the central work-stage
modeled as a cylinder limited by one top plane. Second, the collision
avoidance between the robots using 5 point-to-point distance constraints
\cite[Constraint (23)]{marinho_dynamic_2019} with points defined
around the robots effector and links. Third, the FoV constraint \cite[the second row of Constraint (23)]{koyama_autonomous_2022}.
Lastly, a focal distance constraint implemented as a pair of complimentary
point-to-point distance constraints with safe and forbidden zone direction
\cite[Constraint (17), (18)]{marinho_dynamic_2019} respectively imposed
on points placed on the drill tip and camera optical center.

\subsection{Adaptive controller}

We solve, at each time step to obtain an adaptation signal $\myvec u_{\estimated a}=\left[\myvec u_{\estimated a_{1}}^{T}\,\myvec u_{\estimated a_{2}}^{T}\right]^{T}$,
the following optimization problem \cite{marinho_adaptive_2022}
\begin{align}
\myvec u_{\estimated a}\in\underset{\dot{\estimated{\myvec a}}}{\text{argmin}} & \norm{\mymatrix J_{\estimated y,\estimated a}\dot{\estimated{\myvec a}}+\eta_{a}\error{\myvec y}}_{2}^{2}+\norm{\mymatrix{\Lambda}_{\estimated a}\dot{\estimated{\myvec a}}}_{2}^{2}\label{eq:adaption_law}\\
\text{subject to} & \begin{alignedat}{1}\myvec W_{\estimated a}\dot{\estimated{\myvec a}} & \preceq\myvec w_{\estimated a}\\
\mymatrix B_{\estimated a}\dot{\estimated{\myvec a}} & \preceq\myvec b_{\estimated a}\\
\mymatrix N_{\estimated a}\dot{\estimated{\myvec a}} & =\myvec 0\\
\error{\myvec x}^{T}\mymatrix J_{x,\estimated a}\dot{\estimated{\myvec a}} & \leq0
\end{alignedat}
\nonumber 
\end{align}
where $\eta_{a}\in(0,\infty)$ is the proportional controller gain,
$\mymatrix{\Lambda}_{\estimated a}$is a positive definite diagonal
damping matrix. The constraints defined by $\left(\myvec W_{\estimated a},\myvec w_{\estimated a}\right)$
and $\left(\myvec B_{\estimated a},\myvec b_{\estimated a}\right)$
are the parameter-space analogous of their configuration-space counterparts
defined in (\ref{eq:control_law}). Lastly, the Lyapunov constraint
is written with
\begin{gather*}
\error{\myvec x}\triangleq\left[\begin{array}{ccc}
\sqrt{\beta\alpha}\vector4\left(\error{\quat t}_{1}\right) & \sqrt{\beta\left(1-\alpha\right)}\vector4\left(\error{\quat r}_{1}\right) & \sqrt{1-\beta}\vector4\left(\error{\quat t}_{2}\right)\end{array}\right]^{T}
\end{gather*}
 and 
\begin{gather*}
\mymatrix J_{x,\estimated a}\triangleq\left[\begin{array}{cc}
\sqrt{\beta\alpha}\mymatrix J_{t,1} & \mymatrix 0\\
\sqrt{\beta\left(1-\alpha\right)}\mymatrix J_{r,1} & \mymatrix 0\\
\mymatrix 0 & \sqrt{1-\beta}\mymatrix J_{t,2}
\end{array}\right].
\end{gather*}

In this paper, we propose a suitable measurement model, $\myvec y$
(see (\ref{eq:measured_line})), an estimated measurement model, $\estimated{\myvec y}$
(see (\ref{eq:estimated_line})), a measurement error, $\error{\myvec y}$
(see (\ref{eq:measurement_error})), the measurement-space Jacobian,
$\mymatrix J_{\estimated y,\estimated a}$ (see (\ref{eq:adaptation_jacobian})),
and complementary task-space Jacobian, $\mymatrix N_{\estimated a}$
(see (\ref{eq:complementary_jacobian})), for single-shot camera measurements
where the relation of one point in the image and its task-space counterpart
is known.
\begin{figure}[t]
\begin{centering}
\includegraphics[width=1\columnwidth]{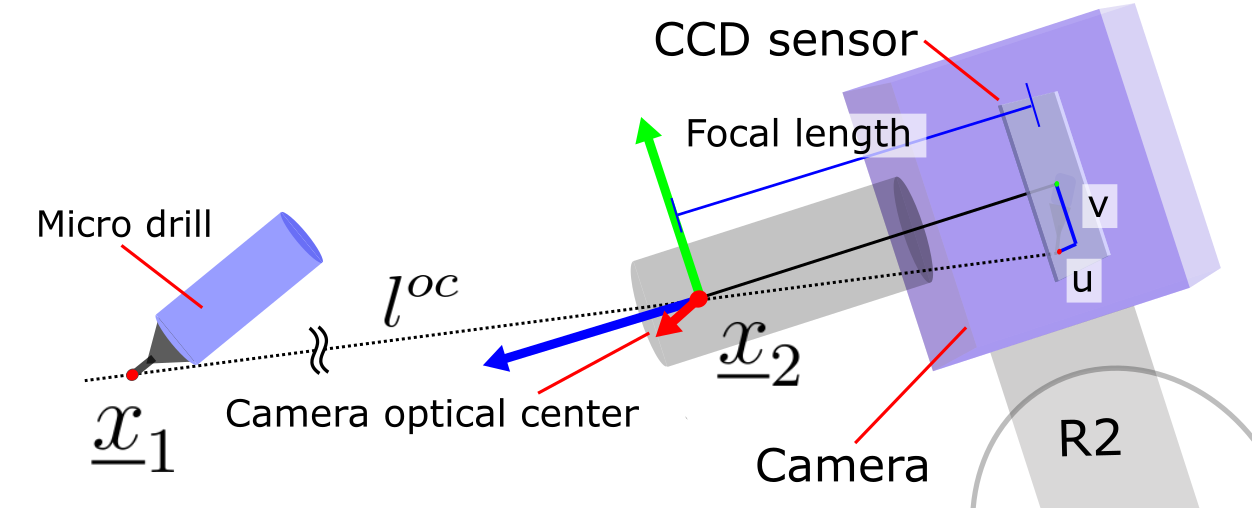}
\par\end{centering}
\caption{\label{fig:camera_model}Relevant elements for the camera extrinsics
modeling of Section~\ref{sec:Proposed-camera-adaptive-methodo}.
Notice that $\protect\quat l^{\text{oc}}\in\mathbb{H}_{p}\cap\mathbb{S}^{3}$
is the task-space measurement, the direction of the line connecting
the pixel $\protect\quat{\rho}^{\text{oc}}\left(u,v\right)\in\mathbb{I}^{2}$
representing the tooltip of $R1$ and the tooltip itself, passing
through the optical center of the camera. Using this information,
we can adapt the estimated robot model and extrinsics to compensate
for initial modeling inaccuracies.}
\end{figure}

\section{Proposed camera-adaptive methodology\label{sec:Proposed-camera-adaptive-methodo}}

This section contains the main technical contribution of this work.
We start by modeling the camera extrinsics in Section~\ref{subsec:Camera-extrinsics-model}
to obtain the measurement space, i.e. model what real-world information
can be obtained from the camera. We then relate that model to a robot-mounted
camera in Section~\ref{subsec:Estimated-space} to obtain the estimated
measurement-space. We then define the measurement error and obtain
the related parametric-space Jacobian in Section~\ref{subsec:Measurement-Error-definition}.
Lastly, we obtain the parametric task-Jacobian projector in Section~\ref{subsec:Parametric-task-Jacobian-project}.

\subsection{Camera extrinsics model: measurement space\label{subsec:Camera-extrinsics-model}}

In this work, we consider the dual-quaternion model for a pin-hole
camera, given as follows. Let a point in world-coordinates $\quat t^{\text{w}}=t_{x}\imi+t_{y}\imj+t_{z}\imk$
represent a un-occluded point of a world object, whose real coordinates
are not measurable directly. The pin-hole camera model is such that
$\quat t^{\text{w}}$, in world coordinates, and its pairing pixel\footnote{Note that the pairing pixel is obtained from the image-processing
algorithm described in Section~(\ref{subsec:U-Net-based-point-tracking}).} $\quat{\rho}^{\text{oc}}\left(u,v\right)=u\imi+v\imj$, in image
coordinates, are connected by a Plucker line that crosses the optical
center, $\quat p_{\text{oc}}^{\text{w}}$, given by
\[
\dq l^{w}\left(u,v,\quat p_{\text{oc}}^{w}\right)\triangleq\quat l^{w}\left(u,v\right)+\dual\left[\quat p_{\text{oc}}^{w}\times\quat l^{w}\left(u,v\right)\right],
\]
where $\quat p_{\text{oc}}^{w}$ is the position of the camera frame,
$\dq x_{\text{oc}}^{w}$, obtained by decomposing the relation $\dq x_{\text{oc}}^{w}=\quat r_{\text{oc}}^{w}+0.5\dual\quat p_{\text{oc}}^{w}\quat r_{\text{oc}}^{w}$,
commonly known as the camera extrinsics. The rotation of the optical
center, $\quat r_{\text{oc}}^{w}$, can be used to map the line direction
between the optical center frame and the world frame $\quat l^{w}\left(u,v\right)=\ad{\quat r_{\text{oc}}^{w}}{\quat l^{\text{oc}}\left(u,v\right)}$.

Under the assumption of the pin-hole camera model, as illustrated
in Fig. \ref{fig:cam_view}, and supposing that the camera constant\footnote{Considering intrinsics adaptation is the topic of ongoing future work.}
\emph{intrinsics} are known\footnote{We used MATLAB's camera calibration \url{https://www.mathworks.com/help/vision/camera-calibration.html}.},
we can find a pixel in task-space, as seen by optical center, as
\[
\quat p_{\rho}^{\text{oc}}\triangleq\quat p_{\rho}^{\text{oc}}\left(u,v\right)=us_{x}\imi+vs_{y}\imj-f\imk,
\]
where $f,s_{x},s_{y}\in\mathbb{R}^{+}-\{0\}$ are the focal length,
the $x-$axis size of the pixel, and the $y-$axis size the each pixel,
respectively. This means that, from a given pixel, we can measure
a line direction, with respect to the optical center, given by
\begin{equation}
\myvec y\triangleq\quat l^{\text{oc}}\left(u,v\right)=\frac{\left(us_{x}\imi+vs_{y}\imj-f\imk\right)}{\norm{\quat p_{\rho}^{\text{oc}}}},\label{eq:measured_line}
\end{equation}
that is not singular even when $u=v=0$ because $f>0$, hence $\norm{\quat p_{\rho}^{\text{oc}}}>0$
$\forall$ $u$, $v$.

\subsection{Robot-mounted camera model: estimated measurement-space\label{subsec:Estimated-space}}

Consider that the camera robot, $R2$, is modeled such that the camera
reference corresponds to its end-effector, i.e.
\[
\estimated{\dq x}_{\text{oc}}^{w}\triangleq\estimated{\dq x}_{2}\left(\myvec q_{2}\right)=\estimated{\quat r}_{2}+\frac{1}{2}\dual\estimated{\quat t}_{2}\estimated{\quat r}_{2}.
\]
With the point of interest on $R1$ defined at the tooltip position
$\estimated{\quat t}_{1}$, obtained from $\estimated{\dq x}_{1}\left(\myvec q_{1}\right)\triangleq\estimated{\dq x}_{1}=\estimated{\myvec r}_{1}+0.5\dual\estimated{\quat t}_{1}\estimated{\myvec r}_{1}$,
we can define the estimated line as
\begin{equation}
\estimated{\quat l}^{\text{oc}}\triangleq\estimated{\quat l}_{2,1}\left(\myvec q\right)=\frac{\estimated{\quat t}_{1}-\estimated{\quat t}_{2}}{\norm{\estimated{\quat t}_{1}-\estimated{\quat t}_{2}}},\label{eq: esimated_line_direction}
\end{equation}
which will never be singular because $\estimated{\quat t}_{2}\neq\estimated{\quat t}_{1}$,
that is, the tip of R1 can never physically be at the optical center
given the collision-avoidance constraints. Lastly, we adjust the reference
frame to be at the optical center, to match our measurement (\ref{eq:measured_line})
\begin{equation}
\estimated{\myvec y}\triangleq\estimated{\quat l}^{\text{oc}}=\ad{\left(\estimated{\quat r}_{2}\right)^{*}}{\estimated{\quat l}_{2,1}}.\label{eq:estimated_line}
\end{equation}
\vspace*{-25pt}

\subsection{Measurement error and adaptation Jacobians\label{subsec:Measurement-Error-definition}}

\begin{figure*}[t]
\centering{}\includegraphics[width=1.8\columnwidth]{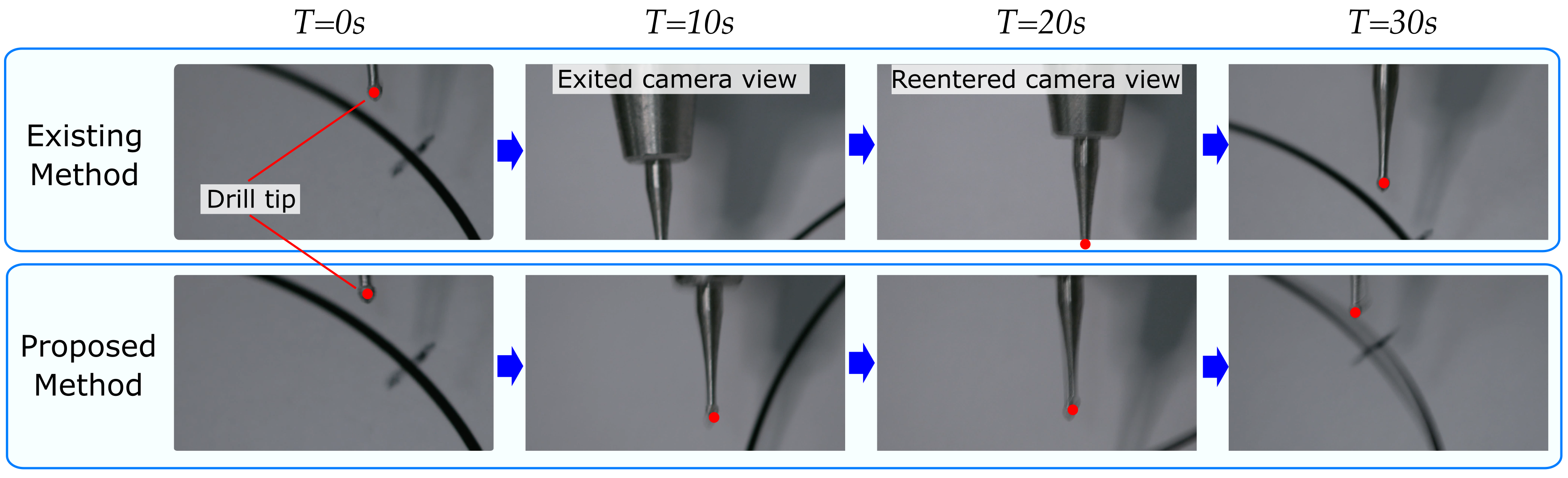}\caption{\label{fig:cam_view}Snapshots of the robot-mounted camera while $R1$
traverses a predefined trajectory. The drilltip frequently leaves
the desired FoV when only kinematic constrained control is used, whereas
it effectively stays within the desired FoV when using the proposed
adaptive strategy.}
\vspace*{-15pt}
\end{figure*}

\begin{figure}
\centering{}\includegraphics[width=0.75\columnwidth]{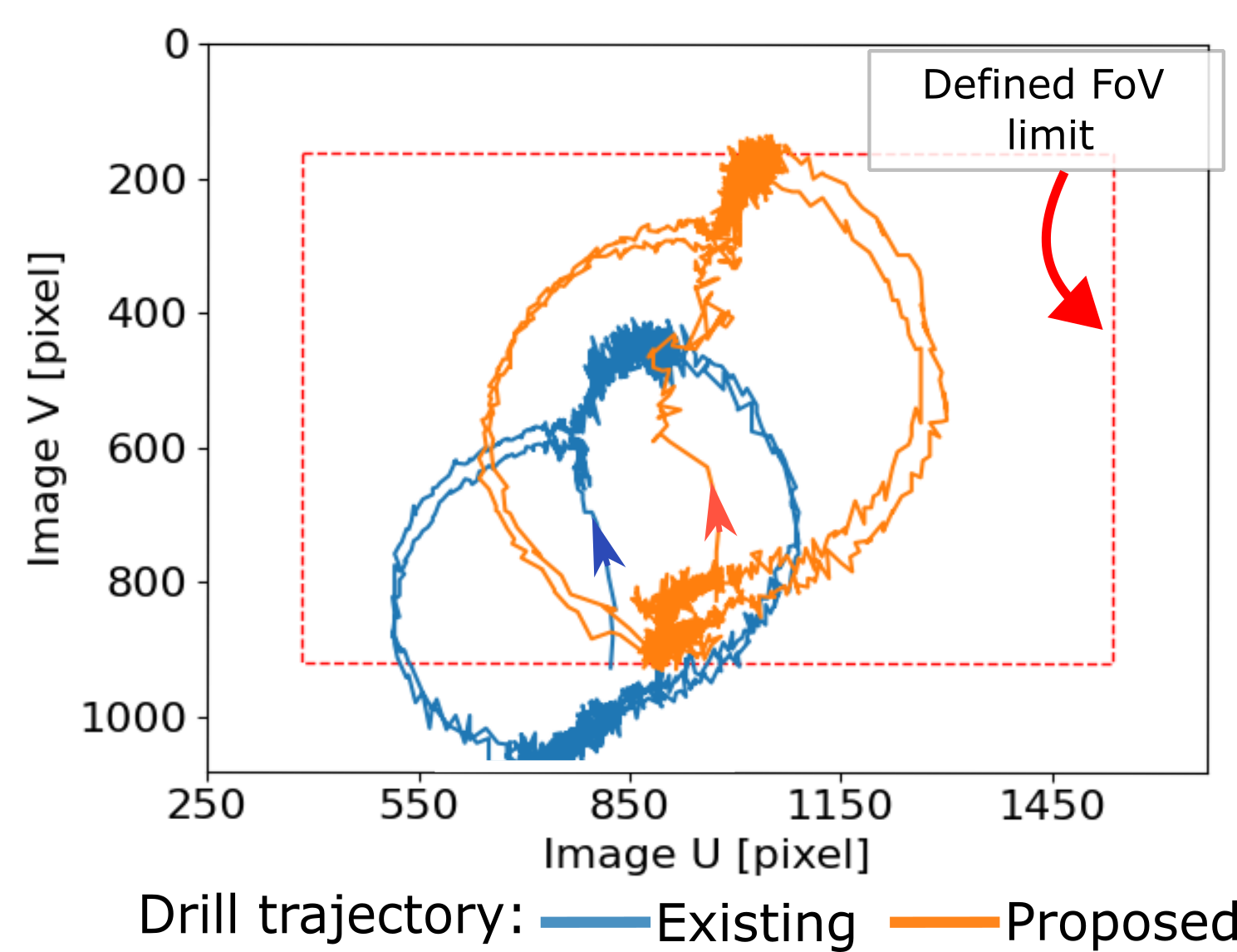}\caption{\label{fig:result_arm_cam}The plot trace the trajectories of the
drill tip under the robot-mounted camera for two rotational period.}
\end{figure}
We can define measurement error as the difference between (\ref{eq:estimated_line})
and (\ref{eq:measured_line}), namely
\begin{equation}
\tilde{\quat y}\triangleq\estimated{\quat l}^{\text{oc}}-\quat l^{\text{oc}}.\label{eq:measurement_error}
\end{equation}

We use the error information to obtain the parameter Jacobian. We
start from the time-derivative of (\ref{eq:estimated_line})
\begin{alignat}{1}
\dot{\estimated{\quat l}}^{\text{oc}} & =\left(\estimated{\quat r}_{\text{2}}\right)^{*}\estimated{\quat l}_{2,1}\estimated{\quat r}_{\text{2}}\label{eq:line_derivative}\\
 & =\left(\dot{\estimated{\quat r}}_{\text{2}}\right)^{*}\estimated{\quat l}_{2,1}\estimated{\quat r}_{\text{2}}+\left(\estimated{\quat r}_{\text{2}}\right)^{*}\dot{\estimated{\quat l}}_{2,1}\estimated{\quat r}_{\text{2}}+\left(\estimated{\quat r}_{2}\right)^{*}\estimated{\quat l}_{2,1}\dot{\estimated{\quat r}}_{\text{2}},\nonumber 
\end{alignat}
meaning that we need to find $\dot{\estimated{\quat r}}_{\text{2}}$
and $\dot{\estimated{\quat l}}_{2}$. The derivative of the rotation
$\dot{\estimated{\quat r}}_{\text{2}}$ is obtained with the rotation
Jacobian, namely
\begin{equation}
\dot{\estimated{\quat r}}_{\text{2}}=\begin{bmatrix}\mymatrix 0 & \mymatrix J_{\estimated r_{2},\estimated a}\end{bmatrix}\dot{\estimated{\myvec a}}.\label{eq:rotation_jacobian}
\end{equation}
As for the line direction derivative, $\dot{\estimated{\quat l}}_{1}$,
we start from
\begin{eqnarray*}
\quat h & \triangleq & \estimated{\quat t}_{1}-\estimated{\quat t}_{2}\implies\\
\dot{\quat h} & = & \dot{\estimated{\quat t}}_{1}-\dot{\estimated{\quat t}}_{2}\implies\\
\vector_{4}\dot{\quat h} & = & \mymatrix J_{\estimated t_{1},\estimated a}\dot{\estimated{\myvec a}}_{1}-\mymatrix J_{\estimated t_{2},\estimated a}\dot{\estimated{\myvec a}}_{2}\\
 & = & \begin{bmatrix}\mymatrix J_{\estimated t_{1},\estimated a} & -\mymatrix J_{\estimated t_{2},\estimated a}\end{bmatrix}\dot{\estimated{\myvec a}}
\end{eqnarray*}
 and substitute it in the derivative of (\ref{eq: esimated_line_direction})
to obtain
\begin{align}
\estimated{\quat l}_{2,1} & =\frac{\quat h}{\norm{\quat h}}=\quat h\norm{\quat h}^{-1}\implies\nonumber \\
\dot{\estimated{\quat l}}_{2,1} & =\dot{\quat h}\norm{\quat h}^{-1}-\quat h\norm{\quat h}^{-2}\dotproduct{\quat h,\dot{\quat h}}\norm{\quat h}^{-1}\implies\nonumber \\
\vector_{4}\dot{\estimated{\quat l}}_{2,1} & =\norm{\quat h}^{-1}\vector_{4}\dot{\quat h}\nonumber \\
 & \overbrace{+\frac{1}{2}\norm{\quat h}^{-3}\hami +_{4}\left(\quat h\right)\left[\hami +_{4}\left(\quat h\right)+\hami -_{4}\left(\quat h\right)\right]}^{\mymatrix A_{1}}\vector_{4}\dot{\quat h}\nonumber \\
 & =\overbrace{\norm{\quat h}^{-1}\begin{bmatrix}\mymatrix J_{\estimated t_{1},\estimated a} & -\mymatrix J_{\estimated t_{2},\estimated a}\end{bmatrix}}^{\mymatrix A_{2}}\dot{\estimated{\myvec a}}\nonumber \\
 & +\underbrace{\mymatrix A_{1}\begin{bmatrix}\mymatrix J_{\estimated t_{1},\estimated a} & -\mymatrix J_{\estimated t_{2},\estimated a}\end{bmatrix}}_{\mymatrix A_{3}}\dot{\estimated{\myvec a}}\nonumber \\
 & =\underbrace{\left(\mymatrix A_{2}+\mymatrix A_{3}\right)}_{\mymatrix J_{\estimated l_{2,1},\estimated a}}\dot{\estimated{\myvec a}}\label{eq:line_jacobian}
\end{align}
Substituting (\ref{eq:rotation_jacobian}) and (\ref{eq:line_jacobian})
in (\ref{eq:line_derivative}) results in
\begin{alignat}{1}
\dot{\estimated{\quat l}}^{\text{oc}} & =\overbrace{\hami -_{4}\left(\estimated{\quat l}_{2,1}\estimated{\quat r}_{\text{2}}\right)\mymatrix C_{4}\begin{bmatrix}\mymatrix 0 & \mymatrix J_{\estimated r_{2},\estimated a}\end{bmatrix}}^{\mymatrix B_{1}}\dot{\estimated{\myvec a}}\nonumber \\
 & +\overbrace{\hami -_{4}\left(\estimated{\quat r}_{\text{2}}\right)\hami +_{4}\left(\estimated{\quat r}_{\text{2}}^{*}\right)\mymatrix J_{\estimated l_{2,1},\estimated a}}^{\mymatrix B_{2}}\dot{\estimated{\myvec a}}\nonumber \\
 & +\overbrace{\hami +_{4}\left(\estimated{\quat r}_{\text{2}}^{*}\estimated{\quat l}_{2,1}\right)\begin{bmatrix}\mymatrix 0 & \mymatrix J_{\estimated r_{2},\estimated a}\end{bmatrix}}^{\mymatrix B_{3}}\dot{\estimated{\myvec a}}\nonumber \\
 & =\underbrace{\left(\mymatrix B_{1}+\mymatrix B_{2}+\mymatrix B_{3}\right)}_{\mymatrix J_{\estimated l_{\text{oc}},\estimated a}}\dot{\estimated{\myvec a}}\nonumber \\
\implies\mymatrix J_{\estimated y,\estimated a} & =\mymatrix B_{1}+\mymatrix B_{2}+\mymatrix B_{3}.\label{eq:adaptation_jacobian}
\end{alignat}

\begin{center}
\vspace*{-30pt}
\par\end{center}

\subsection{Parametric task-Jacobian projector, $\protect\mymatrix N_{\protect\estimated a}$\label{subsec:Parametric-task-Jacobian-project}}

As stated in the adaptation law, the purpose of the parametric task-Jacobian
projector exist to prevent disturbance in the unmeasured variables.
Given the task-space measurement definition, the real translation
of the drill in the world frame $\quat t_{1}$is unknown, therefore
we can fix
\begin{gather*}
\dot{\estimated{\quat t}}\left(\estimated{\myvec a},\myvec q\right)\implies\underbrace{\left[\begin{array}{cc}
\mymatrix J_{\estimated t_{1},\estimated a} & \mymatrix 0\end{array}\right]}_{\mymatrix N_{\estimated a,1}}\dot{\estimated{\myvec a}}=0.
\end{gather*}
Additionally, we cannot know rotations about $\estimated{\quat l}_{2,1}$,
hence we will impose a 1 DoF constraint such that the angular velocity
$\quat{\omega}_{l}$ is perpendicular $\estimated{\quat l}_{2,1}$,
namely
\begin{alignat*}{1}
\dotproduct{\estimated{\quat l}_{2,1},\quat{\omega}_{l}(\estimated a,q)} & =0\\
\implies\left\langle \estimated{\quat l}_{2,1},2\dot{\estimated{\quat r}}_{2}\estimated{\quat r}_{2}^{*}\right\rangle  & =0\\
\underbrace{2\vector_{4}\estimated{\quat l}_{2,1}\,\hami -\left(\estimated{\quat r}_{2}^{*}\right)\begin{bmatrix}\mymatrix 0 & \mymatrix J_{\estimated r_{2},\estimated a}\end{bmatrix}}_{\mymatrix N_{\estimated a,2}}\dot{\estimated{\myvec a}} & =0
\end{alignat*}
\vspace*{-4pt}which are combined into 
\begin{gather}
\mymatrix N_{\estimated a}=\begin{bmatrix}\mymatrix N_{\estimated a,1}^{T} & \mymatrix N_{\estimated a,2}^{T}\end{bmatrix}^{T}\implies\mymatrix N_{\estimated a}\dot{\estimated{\myvec a}}=0.\label{eq:complementary_jacobian}
\end{gather}

\begin{center}
\vspace*{-15pt}
\par\end{center}

\section{Experiments and results}

A set of experiments was conducted to evaluate the effectiveness
of the proposed method with the setup shown in Fig. \ref{fig:sytem}.
We defined a (1150x750 pixels) subregion of the camera as the FoV
as shown in Fig. \ref{fig:cam_view}, so that deviations of the FoV
could be measured. We implemented two controllers. The first controller
represents the existing model-based control strategy, whereas the
second controller uses the proposed adaptive strategy to compensate
for inaccuracies in the parameters of $R2$ arms, including camera
extrinsics.

At the start of the experiment, we manually adjusted the initial pose
of $R1$ at the start of the trajectory and $R2$ camera to point
at the drill with the tip in view and in focus. We also established
parameters for FoV constraint with ,$\theta_{safe}=0.55\lyxmathsym{\textdegree}$
focus distance constraints with $d_{image}=0.405\text{m}$, and adaptation
gain with $\eta_{a}=7$. The experiment involved automatically moving
the drill tip along a pre-defined circular path with a radius r=4
cm and a period of 30s. The path followed by the tooltip of $R1$
in the image of the camera of $R2$ was used to calculate the proportion
of time that the tool remained inside the robot-mounted camera's field.
\begin{table}
\caption{\label{tab:tracking-result}Real FoV constraint maintenance}

\begin{centering}
\scalebox{0.8}{%
\begin{tabular}{|c|c|c|}
\cline{2-3} \cline{3-3} 
\multicolumn{1}{c|}{} & Existing & Proposed\tabularnewline
\hline 
Duration Ratio & 54.4\%\% & 94.1\%\tabularnewline
\hline 
Maximum Deviation ($\theta_{\text{FoV}}-\theta_{safe}$) & $\geq0.68\lyxmathsym{\textdegree}$ {*} & $0.27\lyxmathsym{\textdegree}$\tabularnewline
\hline 
\end{tabular}}
\par\end{centering}
\medskip{}

{\small{}{*}due to the drill tip deviating off the camera view, only
a lower bound value is available.}{\small\par}
\end{table}

The experimental results are presented in Table \ref{tab:tracking-result}
showing the percentage of time that the \emph{real} drill tip remained
within the FoV of the robot-mounted camera and the maximum deviation
outside of the \emph{real} FoV constraint. It is important to notice
that the \emph{estimated} FoV is always kept for both controllers.
Nonetheless, under our proposed method, the system successfully retained
the drill tip within the \emph{real} FoV for 94.1\% of the time, compared
to the 54.4\% achieved by the existing method. The difference is visible
in Fig. \ref{fig:result_arm_cam}, in which we show the trajectory
of the drill tip from the perspective of the robot-mounted camera.
Our proposed method consistently kept the drill tip within the projected
rectangle, representing the FoV constraints. Conversely, the control
group's FoV constraints have an offset pointing slightly above the
actual drill tip, demonstrating a mismatch between the estimated and
actual model of the robots and the need for parameter adaptation.
After improving the frequency of the image-processing pipeline, currently
at 32Hz, and the vibration seen in the robot-mounted camera sourcing
from the robot's joint motion, likely from the weight of the camera
assembly which is near the limit of the robot's payload, we expect
to get better performance using the proposed strategy.

\section{Conclusions}

In this paper, we introduced a framework for automated camera positioning,
employing kinematic constraints to address the camera's field of view
challenges. Simultaneously, we incorporated a parameter adaptation
strategy by integrating keypoint detection into task-space measurements.
Future work should incorporate additional task-space measurements
for improved task performance. There is also an avenue to improve
system functionality by autonomously controlling the neutral point-of-view.

\bibliographystyle{IEEEtran}
\addcontentsline{toc}{section}{\refname}\bibliography{references/cam_automation,references/warning_entry}

\end{document}

%% file: macros/macros.tex
\global\long\def\dq#1{\underline{\bm{#1}}}%

\global\long\def\quat#1{\boldsymbol{#1}}%

\global\long\def\mymatrix#1{\boldsymbol{#1}}%

\global\long\def\myvec#1{\boldsymbol{#1}}%

\global\long\def\mapvec#1{\boldsymbol{#1}}%

\global\long\def\crossproduct#1#2{\frac{#1#2-#2#1}{2}}%

\global\long\def\dualvector#1{\underline{\boldsymbol{#1}}}%

\global\long\def\dual{\varepsilon}%

\global\long\def\dotproduct#1{\langle#1\rangle}%

\global\long\def\norm#1{\left\Vert #1\right\Vert }%

\global\long\def\mydual#1{\underline{#1}}%

\global\long\def\hami#1{\overset{#1}{\operatorname{\mymatrix H}}}%

\global\long\def\hamidq#1#2{\overset{#1}{\operatorname{\mymatrix H}}_{8}\left(#2\right)}%

\global\long\def\hamilton#1#2{\overset{#1}{\operatorname{\mymatrix H}}\left(#2\right)}%

\global\long\def\hamiquat#1#2{\overset{#1}{\operatorname{\mymatrix H}}_{4}\left(#2\right)}%

\global\long\def\tplus{\dq{\mathcal{T}}}%

\global\long\def\gett#1{\dq{\mathcal{T}}\left(#1\right)}%

\global\long\def\dgett#1{\dq{\mathcal{T}}'\left(#1\right)}%

\global\long\def\getp#1{\operatorname{\mathcal{P}}\left(#1\right)}%

\global\long\def\dgetp#1{\operatorname{\mathcal{P}}'\left(#1\right)}%

\global\long\def\getd#1{\operatorname{\mathcal{D}}\left(#1\right)}%

\global\long\def\swap#1{\text{swap}\{#1\}}%

\global\long\def\imi{\hat{\imath}}%

\global\long\def\imj{\hat{\jmath}}%

\global\long\def\imk{\hat{k}}%

\global\long\def\real#1{\operatorname{\mathrm{Re}}\left(#1\right)}%

\global\long\def\imag#1{\operatorname{\mathrm{Im}}\left(#1\right)}%

\global\long\def\imvec{\boldsymbol{\imath}}%

\global\long\def\vector{\operatorname{vec}}%

\global\long\def\mathpzc#1{\fontmathpzc{#1}}%

\global\long\def\cost#1#2{\underset{\text{#2}}{\operatorname{\text{cost}}}\left(\ensuremath{#1}\right)}%

\global\long\def\diag#1{\operatorname{diag}\left(#1\right)}%

\global\long\def\frame#1{\mathcal{F}_{#1}}%

\global\long\def\ad#1#2{\text{Ad}\left(#1\right)#2}%

\global\long\def\adsharp#1#2{\text{Ad}_{\sharp}\left(#1\right)#2}%

\global\long\def\error#1{\tilde{#1}}%

\global\long\def\derror#1{\dot{\tilde{#1}}}%

\global\long\def\dderror#1{\ddot{\tilde{#1}}}%

\global\long\def\spin{\text{Spin}(3)}%

\global\long\def\spinr{\text{Spin}(3){\ltimes}\mathbb{R}^{3}}%

%% file: macros/macros_adaptive.tex
\global\long\def\unitdqspace{\dq{\mathcal{S}}}%

\global\long\def\unitquatspace{\mathbb{S}^{3}}%

\global\long\def\hp{\mathbb{H}_{p}}%

\global\long\def\spin{\text{Spin}(3)}%

\global\long\def\spinr{\text{Spin}(3){\ltimes}\mathbb{R}^{3}}%

\global\long\def\estimated#1{\hat{#1}}%

\global\long\def\jointspace{\mathcal{Q}}%

\global\long\def\parameterspace{\mathcal{A}}%

\global\long\def\taskspace{\mathscr{T}}%

\global\long\def\measurespace{\mathcal{Y}}%

\global\long\def\restrictedset#1{#1_{\mathrm{r}}}%